\documentclass[sigconf, nonacm, authorversion]{acmart}

\usepackage{makecell}
\usepackage{multirow}
\usepackage{bbding}
\usepackage{CJKutf8}
\usepackage{xcolor,soul}

\usepackage{caption}
\AtBeginDocument{%
  }

\begin{document}
\settopmatter{authorsperrow=4}
\title{MSBraM: A Multi-scale Self-supervised Brain Foundation Model for Hierarchical EEG Dynamics Learning}



\author{Tao Zhou}
\email{zhooutao@hnu.edu.cn}
\affiliation{%
  \institution{Hunan University}
  \city{Changsha}
  \country{China}
}

\author{Jing Han}
\email{jhan@hnu.edu.cn}
\affiliation{%
  \institution{Hunan University}
  \city{Changsha}
  \country{China}
}

\author{Lingyu Shu}
\email{shulingyu@hnu.edu.cn}
\affiliation{%
  \institution{Hunan University}
  \city{Changsha}
  \country{China}
}

\author{Zixing Zhang}
\authornote{Zixing Zhang is the corresponding author.}
\email{zixingzhang@hnu.edu.cn}
\affiliation{%
  \institution{Hunan University}
  \city{Changsha}
  \country{China}
}

\renewcommand{\shortauthors}{Tao Zhou, Jing Han, Lingyu Shu, Zixing Zhang.}

\begin{abstract}
Self-supervised foundation models have recently shown strong potential for electroencephalogram (EEG)-based analysis. However, existing approaches struggle to capture the inherently multi-scale temporal structure of EEG signals, where local neural patterns and long-range dependencies jointly encode task-relevant information. This limitation hampers cross-scale representation learning and generalization across diverse downstream tasks.
To address this challenge, we propose \textbf{MSBraM}, a \textbf{M}ulti-\textbf{S}cale self-supervised \textbf{Bra}in foundation \textbf{M}odel designed to learn hierarchical EEG representations. MSBraM follows a two-stage pretraining framework. First, a multi-scale neural tokenizer discretizes raw EEG signals into semantic codes at different temporal resolutions via vector-quantized reconstruction. Second, the model is pretrained to predict masked codes using a curriculum multi-scale masking strategy, progressively integrating fine-grained local patterns with global temporal context.
We pretrain MSBraM on over 2,400 hours of EEG data and evaluate it across 10 downstream tasks on 12 public datasets. Extensive experiments show that MSBraM achieves superior performance on other state-of-the-art pretrained models, demonstrating strong generalization and transferability. These results indicate that explicitly modeling multi-scale temporal dynamics is critical for effective EEG foundation models. 
\end{abstract}

\ccsdesc[500]{Computing methodologies~Artificial intelligence}
\ccsdesc[500]{Applied computing~Life and medical sciences}

\keywords{EEG self-supervised Learning, EEG foundation model, Multi-scale EEG Dynamic Learning}


\maketitle

\section{Introduction}

Electroencephalogram (EEG) signals are recordings of the neural electrical activity in the human brain, which is measured by electrodes placed on the scalp surface. As they reveal the physical state of the brain, EEG signals are widely used for brain function research, disease diagnosis, and brain--computer interface (BCI) applications, such as emotion recognition~\cite{liu2025eeg,pan2025real,liu2024vsgt}, epilepsy detection~\cite{shoeb2009chbmit,zhou2025tyee}, mental disorder diagnosis~\cite{zyma2019eegmat}, fatigue assessment~\cite{zheng2017seedvig}, and gait prediction~\cite{he2018mobi}. Early studies primarily relied on traditional machine learning methods, which required hand-crafted feature engineering based on expert knowledge. With the rapid advancement of deep learning, a variety of deep models have been developed for EEG analysis. For example, EEGNet~\cite{lawhern2018eegnet}, SPaRCNet~\cite{jing2023sparcnet}, ContraWR~\cite{yang2021contrawr} and FFCL~\cite{li2022ffcl} employ convolution neural network, while EEGConformer~\cite{song2022eegconformer}, CNN-Transformer~\cite{peh2022cnntransformer} and ST-Transformer~\cite{song2021sttransformer} introduce Transformer for architectures for EEG signal modeling. 
Despite their considerable success, these methods are often tailored for specific tasks, which limits their generalization and also hinders transfer to new datasets or tasks.

Inspired by the success of self-supervised learning (SSL) in computer vision (CV) and natural language processing (NLP), recent years have witnessed a growing interest in applying these techniques to EEG signals. For instance, self-supervised foundation EEG models such as BIOT~\cite{yang2023biot}, LaBraM~\cite{jiang2024labram}, EEGPT~\cite{wang2024eegpt}, and CBraMod~\cite{wang2025cbramod} have demonstrated strong performance and promising generalization across diverse EEG datasets and tasks. Nevertheless, these advances primarily translate to improved downstream performance without addressing the fundamental limitations in representation learning. Specifically, these methods generally produce low-quality, homogeneous, or task-specific representations, ignoring the hierarchical dynamics inherent to EEG signals.


The multi-scale nature of EEG signals refers to the fact that neural information is simultaneously encoded across dynamically interacting \textit{spatial}, \textit{temporal}, and \textit{spectral} scales. Temporally, transient discharges lasting milliseconds (e.g., epileptic spikes) coexist with slow-wave oscillations that persist for seconds. Spectrally, different cognitive or pathological states are associated with characteristics and coupling patterns in the classical frequency bands (delta, theta, alpha, beta, gamma). Neglecting this complex, multi-scale dynamic inherently limits model performance, resulting in representations that are neither sufficiently rich nor readily transferable to unseen tasks or datasets. For instance, representations effective for event-level detection (e.g., epileptic spike recognition) often generalize poorly to long-horizon state decoding tasks, such as sleep staging or mental state assessment~\cite{shoeb2009chbmit,phan2019seqsleepnet,zyma2019eegmat}.
Therefore, how to explicitly and effectively model and fuse these multi-scale features and interactions in a SSL represents a core challenge, and a key opportunity for building general and powerful foundation models for EEG.

To address this limitation, we propose MSBraM, a multi-scale structured brain foundation model for hierarchical EEG dynamics learning. Following the LaBraM architecture, MSBraM is pre-trained on a large-scale dataset over 2,400 hours via a two-stage pipeline. In the first stage, a multi-scale neural tokenizer discretizes raw EEG signals into semantically rich codes by a multi-scale codebook, effectively capturing multi-scale patterns from fine-grained to coarse-grained resolutions. In the second stage, we introduce a curriculum multi-scale masking strategy to address the varying contextual dependencies inherent in different temporal scales. By dynamically scaling the learning interest, this strategy enables the model to learn from local features to global context gradually. As a result, MSBraM is the first foundation model to systematically integrate multi-scale architecture into a general SSL paradigm for EEG signals. The main contributions of this paper are summarized as follows:

\begin{itemize}
    \item \textbf{Multi-scale architecture.} We introduce MSBraM, a multi-scale architecture designed to capture the inherent multi-scale characteristics in EEG signals, spanning temporal scales from millisecond transients to slow-wave oscillations. 
    
    \item \textbf{Multi-scale neural tokenizer.} We propose a multi-scale tokenizer with a novel multi-scale codebook to discretize raw EEG signals into semantically rich codes at multiple temporal resolutions, which is trained via vector-quantized Fast Fourier Transform (FFT) reconstruction.
    
    \item \textbf{Curriculum multi-scale masking strategy.} We develop a curriculum multi-scale masking strategy that dynamically schedules the masking ratio to address varying contextual dependencies across scales, which enables MSBraM to learn from local patterns to global context gradually and enhances its capacity to capture complex multi-scale dependencies in EEG signals.
    
    \item \textbf{Extensive benchmark evaluation.} We conduct a comprehensive evaluation of MSBraM across 10 tasks and 12 datasets. Experimental results demonstrate that our MSBraM achieves state-of-the-art (SOTA) performance across diverse benchmarks.
\end{itemize}

\section{Related Work}

\textbf{Self-supervised learning for EEG signals.}
Inspired by the success of self-supervised learning in computer vision and natural language processing, researchers have recently begun to explore its application to EEG signal modeling. For instance, BENDR~\cite{kostas2021bendr} utilized the wav2vec 2.0 architecture from speech recognition and employed a contrastive learning paradigm for pre-training, demonstrating the feasibility of transferring SSL paradigms to EEG modeling. Similarly, BrainBERT~\cite{wang2023brainbert} randomly masked portions of stereo EEG signals and predicted the masked representations based on contextual information. Building on this, Brain~\cite{zhang2023brant} and Brain-2~\cite{yuan2402brainwave} performed masked modeling on a large clinical intracranial neural signal dataset, achieving performance improvement across multiple downstream tasks. Furthermore, Brain-X~\cite{zhang2024brantx} captured multi-modal information to enhance model performance by aligning EEG signals with other modalities of physiological data. 

Subsequently, BIOT~\cite{yang2023biot} introduced the Biosignal Transformer and contrastive learning to build a general-purpose foundation model. In contrast to raw signals reconstruction, EEG2Rep~\cite{navid2024eeg2rep} employed a self-prediction approach that targets high-level representations. More recently, LaBraM~\cite{jiang2024labram} proposed an EEG neural tokenizer to enable masked EEG modeling, achieving state-of-the-art results on several EEG datasets. EEGPT~\cite{wang2024eegpt} further combined masked modeling with EEG representation reconstruction to improve performance. Additionally, to address the heterogeneity of EEG montages, MMM~\cite{yi2023mmm} employed geometry-aware modeling to learn montage-agnostic representations. Meanwhile, CBraMod~\cite{wang2025cbramod} proposed a Criss-Cross Attention mechanism to capture temporal and spatial representations for multi-channel EEG data simultaneously.
In this study, we systematically evaluate and compare the proposed MSBraM with the aforementioned SSL EEG models, and MSBraM achieves superior performance across diverse tasks and datasets.

\textbf{Multi-scale Model.}
Multi-scale architecture serves as the building blocks in deep learning for capturing patterns across different spatial and temporal resolutions. U-Net~\cite{olaf2015unet} introduced a novel multi-scale fusion paradigm through its encoder-decoder and skip connections, which links high-resolution details with contextual semantics. This design has become a standard in biomedical image analysis. Subsequently, the Feature Pyramid Network (FPN)~\cite{yi2017fpn} augmented a standard backbone with a top-down pathway for multi-scale representation. PANet~\cite{liu2018panet} and BiFPN~\cite{tan2020bifpn} further enhanced feature aggregation by incorporating bidirectional pathways. While NAS-FPN~\cite{ghiasi2019nasfpn} automated the design of cross-scale connections through neural architecture search.
Additionally, HRNet~\cite{wang2021hrnet} pioneered a distinct approach by maintaining high-resolution representations throughout the network, enabling continuous multi-scale fusion and achieving strong performance in dense prediction tasks such as human pose estimation. Building on this, HRFormer~\cite{yuan2021hrformer} integrated transformer modules into the high-resolution framework to capture long-range dependencies, while UHRNet~\cite{wang2022uhrnet} combined HRNet with U-Net to further strengthen multi-scale representation learning. Besides, CEDNet~\cite{zhang2025cednet} extended these principles and drove feature learning across stages by incorporating high-level semantic information into earlier coarse-grained representations.

Despite the success of multi-scale paradigms in computer vision, their systematic application in EEG foundation models remains underexplored. To bridge this gap, we introduce MSBraM, a multi-scale architecture tailored to capture the inherent multi-scale dynamics in EEG signals.
\section{Methodology}

\begin{figure*}[!ht]
    \centering
    \includegraphics[width=\linewidth]{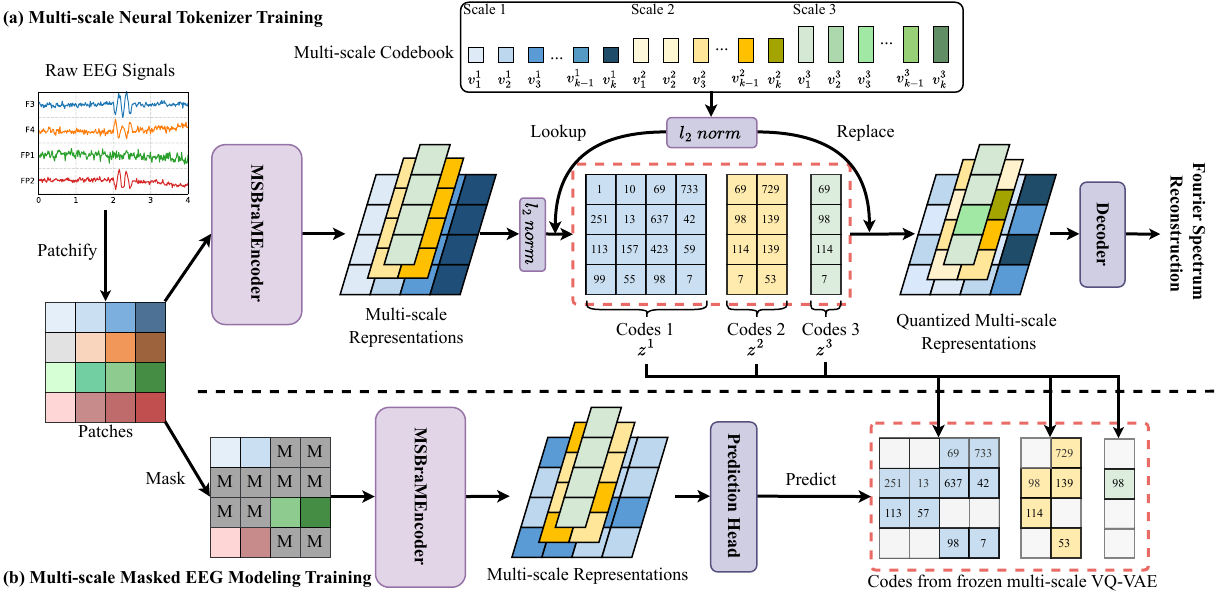}
    \caption{The pretraining pipeline of MSBraM. It contains two stages: (a) \textbf{Multi-scale Neural Tokenizer Training}, i.e., the multi-scale tokenizer learns scale-specific codebooks by reconstructing the Fourier spectrum, discretizing signals into tokens across temporal resolutions; and (b) \textbf{Multi-scale Masked EEG Modeling Training}, i.e., the encoder is pre-trained via masked prediction, driven by our curriculum multi-scale masking strategy, which dynamically schedules ratios to learn hierarchical representations.}
    \label{fig:flow}
\end{figure*}

\begin{figure}[!ht]
    \centering
    \includegraphics[width=\linewidth]{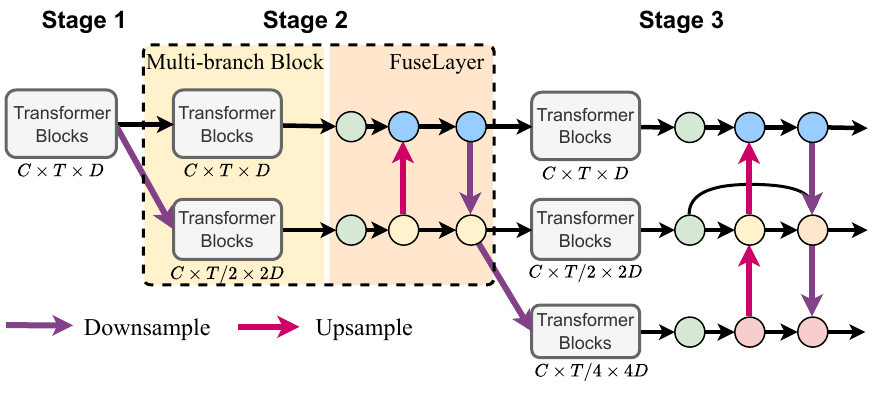}
    \caption{Architecture of MSBraMEncoder. It comprises several stacked stages, with each stage utilizing a \textit{downsample module} to produce a coarser scale, a \textit{multi-branch Transformer} for representation modeling, and a \textit{fusion module} (FuseLayer) for cross-scale fusion.}
    \label{fig:architecture}
\end{figure}

In this section, we introduce the MSBraM architecture and its two-stage pretraining pipeline, following the paradigm of LaBraM. As shown in Figure~\ref{fig:flow}, the first stage trains a multi-scale neural tokenizer with multi-scale codebooks to discretize raw EEG signals into semantic patches at different temporal scales. While the second stage conducts masked EEG modeling to learn contextual representations by predicting masked patches across scales. The details of each component are presented in the following subsections.

\subsection{Model Architecture}
\label{sec:architecture}

\textbf{Patching, Patch Encoder and Positional Encoding} Given an EEG input signal $\mathcal{S} \in \mathbb{R}^{C\times T}$, where C is the number of electrodes and T is the number of timestamps, we first segment it into a set of non-overlapping patches $\mathcal{X} \in \mathbb{R}^{C \times N \times P}$.  Here, $P$ denotes the patch size, $N$ is the number of patches such that $T=N \times P$. These patches 
$\mathcal{X}$ are then fed into a lightweight PatchEncoder to capture local features. Specifically, the encoder consists of three sequential blocks, each containing a 1-D convolution layer, followed by group normalization and a GELU activation function. 
To capture both the temporal and spatial dependencies, we follow the design of LaBraM and introduce learnable temporal embeddings $\{e^t_1, e^t_2, ..., e^t_{T_{max}}\}$ and spatial positional embeddings $\{e^c_1, e^c_2, ..., e^c_C\}$, where $T_{max} > N$ denotes the maximum sequence length.  For each patch located at temporal index $i$ and electrode index $j$, its final representation $\tilde x_{i,j} \in \mathbb{R}^{C \times N \times D}$ is obtained as:
\begin{equation}
\tilde x_{i,j} = \text{PatchEncoder}(x_{i,j}) + e^t_i + e^c_j,
\end{equation}
where $e^t_i$ and $e^c_j$ are retrieved by indexing the corresponding temporal and spatial positional embedding tables and are broadcast-added across the patch sequence.

\textbf{MSBraMEncoder} To capture multi-scale EEG representations, the patches are fed into the MSBraMEncoder, which is composed of multiple stacked stages. As shown in Figure~\ref{fig:architecture}, each stage follows a ``downsampling – multi-branch Transformer encoding – representation fusion'' paradigm and consists of the following three components:

\textit{Downsampling Module} To capture spatial and temporal dependencies at a coarser temporal scale, each stage begins with a downsampling module. This module is applied solely to the output of the last (coarsest) branch from the previous stage, and produces a new candidate branch with a lower temporal resolution. Specifically, the downsampling module is implemented via a PatchMerging layer. It first concatenates each group of two consecutive patches in the sequence, and then applies a linear layer to transform these features into a new representation at the coarser resolution. 
Formally, given the output $H^l=\{h^l_1, h^l_2, ..., h^l_s\}$ of the $l$-th stage, where s is the number of scales, and $h^l_s \in \mathbb{R}^{C \times N_s \times D_s}$ is the representation of the coarsest scale. The downsampling operation on $h^l_s$ is defined as:
\begin{equation}
h^l_{s+1} = \text{Linear}(\text{Concat}(h^l_{s,[1:2]}, h^l_{s,[3:4]}, ...)),
\end{equation}
where $h^l_{s+1} \in \mathbb{R}^{\frac{N_s}{2} \times 2D}$. Finally, the output representation of the downsampling module is $H^{l+1} = \{ h^l_1, h^l_2, ..., h^l_s,h^l_{s+1}\}$.

\textit{Multi-branch Transformer Block} To capture temporal and spatial information across different scales, we utilize a multi-branch TransformerEncoder module, which consists of a set of parallel Transformer encoders. Each branch comprises n transformer encoder layers and learns from the patches at the corresponding scale. 
Given input containing $K$ branches, denoted as $\{h^{l-1}_k\}_{k=1}^K$, and for the $k$-th branch, the representation is performed independently as follows:
\begin{equation}
h^l_{k} = \text{TransformerEncoder}(h^{l-1}_k), \forall k \in \{1,..., K\}.
\end{equation}

Following the same design as LaBraM, we flatten both the channel (C) and length (L) dimensions into a 1-D sequence, which allows MSBraM to jointly model temporal and spatial dependencies in a single, coherent representation space.

\textit{Fusion Module} In contrast to the simple additive or concatenative fusion in HRNet, we employ a Bidirectional Feature Pyramid Network (BiFPN) to adaptively integrate multi-scale representations. BiFPN utilizes a learnable weighted fusion mechanism that dynamically emphasizes the most relevant features from each scale, which is important for modeling the diverse spatio-temporal dynamics in EEG signals. In more detail, the BiFPN module $\mathcal{F}_{\text{bifpn}}(\cdot)$ fuses the outputs $\{h_k^l\}_{k=1}^K$ from the multi-branch Transformer modules to produce a unified and enhanced set of multi-scale representations:
\begin{equation}
\{\tilde h^l_k\}_{k=1}^K=\mathcal{F}_{\text{bifpn}} (\{ h^l_k\}_{k=1}^K),
\end{equation}
where $\mathcal{F}_{\text{bifpn}}$ performs bidirectional pathways multi-scale fusion, and latent representations at each scale are added via normalized learnable weights $w_i$.

In summary, each stage of the MSBraMEncoder follows a defined three-step workflow to extract and refine multi-scale representations. First, the Downsampling Module introduces a new, coarser scale into the hierarchy. Subsequently, the Multi-branch Transformer Block processes all scales in parallel to model temporal and spatial dependencies within each resolution. Finally, the Fusion Module (BiFPN) integrates information across scales via learnable weighted combinations, yielding a coherent multi-scale feature set for the stage.  By stacking such stages, we build the complete MSBraMEncoder, which operates in a fine-to-coarse manner to learn multi-scale EEG representations that effectively capture both fine-grained details and long-range contextual patterns.

\subsection{Multi-scale Neural Tokenizer}
Differing from LaBraM, which employs a single codebook, we propose a multi-scale neural tokenizer to learn discrete latent representations of EEG signals at multiple temporal resolutions. As shown in Figure~\ref{fig:flow}, the tokenizer consists of two core designs: the multi-scale model architecture (Section~\ref{sec:architecture}) and a novel multi-scale codebook. The design of the multi-scale codebook is detailed next.

\textbf{Multi-scale Codebook} The multi-scale codebook aims to learn discriminative prototypes for patterns at different temporal scales. Specifically, we construct a set of codebooks $\mathcal{V} = \{\mathcal{V}^1, \mathcal{V}^2, ..., \mathcal{V}^K\}$, where $K$ is the number of scales. Each $\mathcal{V}^k \in \mathbb{R}^{V \times D}$ contains $V$ prototype vectors of dimension $D$. Given latent representation from the $k$-th scale of the MSBraMEncoder, we lookup nearest neighbor in the corresponding codebook $\mathcal{V}^k$ via cosine similarity:
\begin{equation}
q^k_{i,j} = \mathop{argmin}_{v^k_m \in \mathcal{V}^k} \Vert l_2 (h^k_{i,j}) - l_2 (v^k_m) \Vert^2,
\end{equation}
where $h^k_{i,j}$ is the latent representation of the $i$-th patch from the $j$-th channel at the $k$-th scale, $q^k_{i,j}$ is the quantized codebook vector. Similar to LaBraM, we utilize the $l_2$ normalization to improve the codebook usage.

\textbf{Loss} After multi-scale vector quantization, the vectors $q^k$ are fed into decoders to reconstruct the Fast Fourier Transform (FFT) spectrum of the raw EEG signal. In contrast to the symmetric encoder-decoder architecture of LaBraM, our decoder employs an asymmetric design, structured as three stacked lightweight 1-D convolutional blocks following the ConvNeXt paradigm~\cite{liu2022convnet}. 
The decoded latent representations are then fed into two separate prediction heads to regress the FFT magnitude 
$o^k_A$ and phase $o^k_{\phi}$ of the original EEG signal, respectively. Thus, the overall reconstruction loss for training the multi-scale vector-quantized neural tokenizer is defined as:
\begin{align}
\nonumber
\mathcal{L}_{recon} = \mathbb{E}_{x \in D} \sum_{k=1}^K \sum_{j=1}^C \sum_{i=1}^N \Vert o^{A,k}_{i,j} - A^k_{i,j} \Vert^2_2 + \Vert o^{\phi, k}_{i,j} - \phi^k_{i,j} \Vert^2_2 \\
+ \Vert sg(l_2 (h^k_{i,j})) - l_2(q^k_{i,j}) \Vert^2_2 + \Vert l_2 (h^k_{i,j}) - sg( l_2(q^k_{i,j})) \Vert^2_2,
\end{align}
where $A^k_{i,j}$ and $\phi^k_{i,j}$ denotes the FFT targets corresponding to the $i$-th patch and $j$-th channel at the $k$-th scale.

\subsection{Multi-scale Masked EEG Modeling}

\textbf{Curriculum Multi-scale Masking} 
To avoid potential information leakage across different scales that allow the model to easily infer masked content, we implement a spatially aligned masking strategy. This strategy produces a mask at the coarsest resolution and projects it to all finer scales, ensuring consistency across the scale space. Consequently, if a region is masked at a fine scale, its corresponding contextual regions at every coarser scale are also masked.
Specifically, given input patches $\tilde{\mathbf{x}} \in \mathbb{R}^{C \times N \times D}$, let $s_{\text{max}}$ be the maximum temporal stride. We first downsample the temporal dimension by $s_{\text{max}}$ to obtain the coarsest grid of length $N_K = \lfloor N / s_{\text{max}} \rfloor$. A binary mask $\mathcal{M} = \{m_{i,j} | m_{i,j} \in \{0,1\}, i \in [1, C], j \in [1, N_K]\}$ is then randomly generated on this grid. This mask is then expanded to match the native resolution of any other scale $k$ with stride $s_k$. Formally, for the $k$-th scale with sequence length $N_k = \lfloor N / s_k \rfloor$, the expanded mask $\mathcal{M}^k \in \mathbb{R}^{C \times N_k}$ is given by
\begin{equation}
\mathcal{M}^k[:, j] = \mathcal{M}[:, \lfloor \frac{ j * s_k}{s_\text{max}} \rfloor], \forall j \in [0, N_k - 1].
\end{equation}
The masked patches are replaced by a learnable mask token vector $m \in \mathbb{R}^D$. Formally, given the binary mask $\mathcal{M}^1$ at the finest scale, the masked input is denoted as $\tilde x^M = \{\tilde x_{i,j}: m_{i,j}=0|i \in [1, C], j \in [1, N]\} \cup \{m: m_{i,j} =1| | i \in [1, C], j \in [1, N] \}$. This masking is only applied at the finest-scale input. The mask and unmasked patches are then propagated through the downsampling operations to coarser scales, ensuring consistent masking.

Considering that different temporal scales involve varying degrees of contextual dependency. A fixed masking strategy (FixedMasking) may therefore limit representation learning. We utilize a curriculum multi-scale masking (CurrMasking) paradigm, starting with a low masking ratio to capture local patterns and gradually increasing it to encourage learning global temporal dependencies. This paradigm enables the model to capture both fine-grained and coarse-grained structures, enhancing its multi-scale representational capacity.
In this paper, the global masking ratio $r_{t}$ at training epoch t is determined by the following schedule:
\begin{equation}
r_t = min(r_{max}, max(r_0, (r_{max} - r_0) * \frac{t - t_0}{t_{max} - t_0})),
\end{equation}
where $r_0$ and $r_{max}$ are the initial and maximum masking ratios, while $t_0$ is the warm‑up epoch, and $t_{max}$ is the total training epoch.

\textbf{Loss} 
The objective of multi-scale masked EEG modeling is to predict discrete patch codes at masked positions across all scales, with prediction targets provided as pseudo-labels by a frozen multi-scale tokenizer.
Formally, let $\mathcal{M}_k$ be the set of masked indices at the $k$-th scale. Given the encoded representations $H^k$ from the MSBraM encoder, a linear head is applied to predict the representation at these masked positions to logits over the patch vocabulary.
The overall loss is the sum of cross-entropy (CE) losses over all masked positions across all scales:
\begin{equation}
    \mathcal{L}_{mem} = \sum_{k=1}^K \sum_{(i,j) \in \mathcal{M}^k} \text{CE}(\text{Linear}(h^k_{i,j}), z^k_{i,j}),
\end{equation}
where $K$ is the number of scales, $h^k_{i,j}$ is the latent representation from $H^k$ at the $i$-th temporal patch and $j$-th channel, and $z^k_{i,j}$ is the code indices obtained from the frozen multi-scale neural tokenizer.

\section{Experiments and Results}

\begin{table*}[t]
    \centering
    \fontsize{9}{10}\selectfont
    \begin{tabular}{lcrrrrr}
        \toprule
        \bf{Tasks} & \bf{Datasets} & \bf{Rate (Hz)} & \bf{\# Channels} & \bf{\# Samples} & \bf{Duration (s)} & \bf{Target} \\
        \midrule
        Event Type Classification   & TUEV{\tiny~\cite{obeid2016tuheeg}}           & 250   & 23 & 112,237 & 5  & 6-class \\
        Abnormal Detection          & TUAB{\tiny~\cite{obeid2016tuheeg}}           & 250   & 23 & 409,083 & 10 & 2-class \\
        \multirow{2}{*}{Motor Imagery Classifcation}
                                    & BCIC-2a{\tiny~\cite{tangermann2012review}}   & 250   & 22 & 5,088   & 4  & 4-class \\
                                    & PhysioNet-MI{\tiny~\cite{schalk2004bci2000}} & 160   & 64 & 9,837   & 4  & 4-class \\
        \multirow{2}{*}{Emotion Recognition}
                                    & SEED-V{\tiny~\cite{liu2021seedv}}            & 1,000 & 62 & 117,744 & 1  & 5-class \\
                                    & FACED{\tiny~\cite{chen2023faced}}            & 250   & 32 & 10,332  & 10 & 9-class \\
        Error Related Negativity    & KaggleERN{\tiny~\cite{margaux2012kaggleern}} & 200   & 56 & 8,840 & 2  & 2-class \\
        Seizure Detection           & CHB-MIT{\tiny~\cite{shoeb2009chbmit}}        & 256   & 16 & 320,848 & 10 & 2-class \\
        Mental Disorder Diagnosis   & Mumtaz2016{\tiny~\cite{wajid2016mumtaz}}     & 256   & 19 & 7,143   & 5  & 2-class \\
        Mental Stress Detection     & EEGMAT{\tiny~\cite{zyma2019eegmat}}          & 500   & 20 & 1,707   & 5  & 2-class \\
        \midrule
        Vigilance Estimation        & SEED-VIG{\tiny~\cite{zheng2017seedvig}}      & 200   & 17 & 20,355  & 8  & regression \\
        Gait Prediction             & MoBI{\tiny~\cite{he2018mobi}}                & 100   & 60 & 57,384  & 2  & regression \\
        \bottomrule
    \end{tabular}
    \caption{Overview of 12 datasets used for 10 downstream tasks, including eight classification tasks and two regression tasks. Statistics include sampling rate, number of channels, sample sizes, sample duration, and target.}
    \label{tab:datasets}
\end{table*}

\begin{figure}[t]
    \centering
    \includegraphics[width=\linewidth]{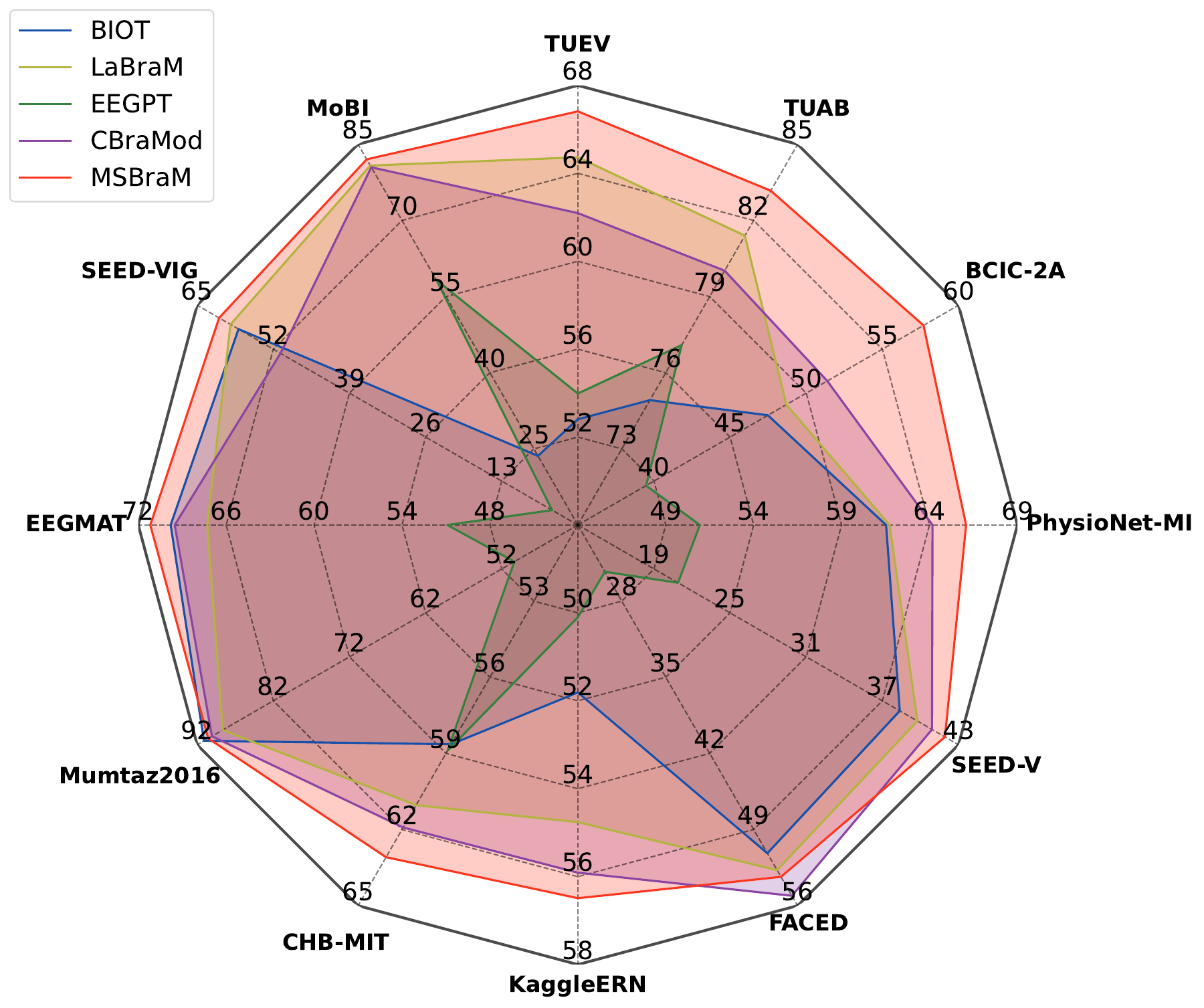}
    \caption{Radar plots comparing MSBraM (red) with other latest foundation models across 10 tasks and 12 datasets in terms of balanced accuracy for classification tasks and Pearson's correlation for regression tasks.}
    \label{fig:radar}
\end{figure}

\begin{table*}[t]
    \centering
    \fontsize{8.5}{10}\selectfont
    \begin{tabular}{lcccccc}
        \toprule
        \multirow{2}{*}{\bf{Methods}}
            & \multicolumn{3}{c}{\bf{TUEV}} & \multicolumn{3}{c}{\bf{TUAB}} \\
            & \bf{Balanced Accuracy} & \bf{Cohen's Kappa} & \bf{Weighted F1}
            & \bf{Balanced Accuracy} & \bf{AUCPR} & \bf{AUROC} \\
        \midrule
        SPaRCNet        & .4161 $\pm$ .0262 & .4233 $\pm$ .0181 & .7024 $\pm$ .0104 & .7896 $\pm$ .0018 & .8414 $\pm$ .0018 & .8676 $\pm$ .0012 \\
        ContraWR        & .4384 $\pm$ .0349 & .3912 $\pm$ .0237 & .6893 $\pm$ .0136 & .7746 $\pm$ .0041 & .8421 $\pm$ .0104 & .8456 $\pm$ .0074 \\
        CNN-Transformer & .4087 $\pm$ .0161 & .3815 $\pm$ .0134 & .6854 $\pm$ .0293 & .7777 $\pm$ .0022 & .8433 $\pm$ .0039 & .8461 $\pm$ .0013 \\
        FFCL            & .3979 $\pm$ .0104 & .3732 $\pm$ .0188 & .6783 $\pm$ .0120 & .7848 $\pm$ .0038 & .8448 $\pm$ .0065 & .8569 $\pm$ .0051 \\
        ST-Transformer  & .3984 $\pm$ .0228 & .3765 $\pm$ .0306 & .6823 $\pm$ .0190 & .7966 $\pm$ .0023 & .8521 $\pm$ .0026 & .8707 $\pm$ .0019 \\
        \midrule
        BIOT            & .5281 $\pm$ .0225 & .5273 $\pm$ .0249 & .7492 $\pm$ .0082 & .7959 $\pm$ .0057 & .8792 $\pm$ .0023 & .8815 $\pm$ .0043 \\
        LaBraM-base     & .6473 $\pm$ .0072 & .6367 $\pm$ .0161 & .8219 $\pm$ .0077 & .8140 $\pm$ .0019 & .8965 $\pm$ .0016 & .9022 $\pm$ .0009 \\
        EEGPT-large     & .5398 $\pm$ .0317 & .6107 $\pm$ .0225 & .7994 $\pm$ .0085 & .7709 $\pm$ .0186 & .8575 $\pm$ .0129 & .8767 $\pm$ .0063 \\
        CBraMod-small   & .6219 $\pm$ .0093 & .5994 $\pm$ .0168 & .7881 $\pm$ .0076 & .8002 $\pm$ .0036 & .8888 $\pm$ .0073 & .8847 $\pm$ .0080 \\
        \midrule
        MSBraM          & \textbf{.6682 $\pm$ .0198} & \textbf{.6785 $\pm$ .0106} & \textbf{.8399 $\pm$ .0054} & \textbf{.8317 $\pm$ .0051} & \textbf{.9034 $\pm$ .0052} & \textbf{.9085 $\pm$ .0024} \\
        \bottomrule
    \end{tabular}
    \caption{Performance comparison between MSBraM and other supervised and self-supervised models on the \textit{event type classification} (TUEV) and the \textit{abnormal detection} (TUAB).}
    \label{tab:tuh}
\end{table*}

\begin{table*}[!t]
    \centering
    \fontsize{8.5}{10}\selectfont
    \begin{tabular}{lcccccc}
        \toprule
        \multirow{2}{*}{\bf{Methods}}
            & \multicolumn{3}{c}{\bf{BCIC-2a}} & \multicolumn{3}{c}{\bf{PhysioNet-MI}} \\
            & \bf{Balanced Accuracy} & \bf{Cohen's Kappa} & \bf{Weighted F1}
            & \bf{Balanced Accuracy} & \bf{Cohen's Kappa} & \bf{Weighted F1} \\
        \midrule
        SPaRCNet        & .4635 $\pm$ .0117 & .2847 $\pm$ .0147 & .4432 $\pm$ .0126 & .5932 $\pm$ .0152 & .4564 $\pm$ .0234 & .5937 $\pm$ .0147 \\
        ContraWR        & .4678 $\pm$ .0125 & .2905 $\pm$ .0160 & .4413 $\pm$ .0142 & .5892 $\pm$ .0133 & .4527 $\pm$ .0248 & .5918 $\pm$ .0116 \\
        CNN-Transformer & .4600 $\pm$ .0108 & .2800 $\pm$ .0148 & .4460 $\pm$ .0114 & .6053 $\pm$ .0118 & .4725 $\pm$ .0223 & .6041 $\pm$ .0105 \\
        FFCL            & .4470 $\pm$ .0143 & .2627 $\pm$ .0176 & .4238 $\pm$ .0139 & .5726 $\pm$ .0092 & .4323 $\pm$ .0182 & .5701 $\pm$ .0079 \\
        ST-Transformer  & .4575 $\pm$ .0145 & .2733 $\pm$ .0198 & .4471 $\pm$ .0142 & .6035 $\pm$ .0081 & .4712 $\pm$ .0199 & .6053 $\pm$ .0075 \\
        \midrule
        BIOT            & .4748 $\pm$ .0093 & .2997 $\pm$ .0139 & .4607 $\pm$ .0125 & .6153 $\pm$ .0154 & .4875 $\pm$ .0272 & .6158 $\pm$ .0197 \\
        LaBraM-base     & .4869 $\pm$ .0085 & .3159 $\pm$ .0154 & .4758 $\pm$ .0103 & .6173 $\pm$ .0122 & .4912 $\pm$ .0192 & .6177 $\pm$ .0141 \\
        EEGPT-large     & .3948 $\pm$ .0266 & .1931 $\pm$ .0354 & .3663 $\pm$ .0349 & .5094 $\pm$ .0108 & .3456 $\pm$ .0144 & .4937 $\pm$ .0121 \\
        CBraMod-small   & .5138 $\pm$ .0066 & .3518 $\pm$ .0094 & .4984 $\pm$ .0085 & .6417 $\pm$ .0091 & .5222 $\pm$ .0169 & .6427 $\pm$ .0100 \\
        \midrule
        MSBraM          & \textbf{.5770 $\pm$ .0070} & \textbf{.4360 $\pm$ .0093} & \textbf{.5673 $\pm$ .0061} & \textbf{.6608 $\pm$ .0040} & \textbf{.5477 $\pm$ .0054} & \textbf{.6615 $\pm$ .0040} \\
        \bottomrule
    \end{tabular}
    \caption{Performance comparison between MSBraM and supervised and self-supervised models on the \textit{motor imagery classification} task ({BCIC-2a} and {PhysioNet-MI}).}
    \label{tab:mi}
\end{table*}

\begin{table*}[!ht]
    \centering
    \fontsize{9}{10}\selectfont
    \begin{tabular}{lcccccc}
        \toprule
        \multirow{2}{*}{\bf{Methods}}
            & \multicolumn{3}{c}{\bf{SEED-VIG}} & \multicolumn{3}{c}{\bf{MoBI}} \\
            & \bf{r} & \bf{R2 Score} & \bf{RMSE $\downarrow$}
            & \bf{r} & \bf{R2 Score} & \bf{RMSE $\downarrow$} \\
        \midrule
        SPaRCNet        & .5715 $\pm$ .0163 & .2433 $\pm$ .0055  & .2798 $\pm$ .0043 & .6466 $\pm$ .0069 & .3688 $\pm$ .0101 & .1151 $\pm$ .0011 \\
        ContraWR        & .5854 $\pm$ .0142 & .2453 $\pm$ .0062  & .2782 $\pm$ .0056 & .1306 $\pm$ .0085 & .0198 $\pm$ .0031 & .1416 $\pm$ .0001 \\
        CNN-Transformer & .5714 $\pm$ .0172 & .2371 $\pm$ .0052  & .2805 $\pm$ .0039 & .3053 $\pm$ .0038 & .1142 $\pm$ .0044 & .1386 $\pm$ .0001 \\
        FFCL            & .5647 $\pm$ .0097 & .2301 $\pm$ .0035  & .2914 $\pm$ .0052 & .4200 $\pm$ .0065 & .1702 $\pm$ .0057 & .1315 $\pm$ .0004 \\
        ST-Transformer  & .5752 $\pm$ .0127 & .2366 $\pm$ .0071  & .2838 $\pm$ .0036 & .7379 $\pm$ .0033 & .4680 $\pm$ .0090 & .1044 $\pm$ .0009 \\
        \midrule
        BIOT            & .5794 $\pm$ .0235 & .1889 $\pm$ .0290  & .2860 $\pm$ .0051 & .2361 $\pm$ .0036 & .0633 $\pm$ .0033 & .1404 $\pm$ .0002 \\
        LaBraM-base     & .5922 $\pm$ .0244 & .2115 $\pm$ .0239  & .2820 $\pm$ .0043 & .8081 $\pm$ .0346 & .6496 $\pm$ .0558 & .0822 $\pm$ .0063 \\
        EEGPT-large     & .5410 $\pm$ .0473 & .0342 $\pm$ .1278  & .3114 $\pm$ .0210 & .5826 $\pm$ .0381 & .3297 $\pm$ .0452 & .1164 $\pm$ .0040 \\
        CBraMod-small   & .5077 $\pm$ .0403 & .1701 $\pm$ .0401  & .2892 $\pm$ .0070 & .8047 $\pm$ .0006 & .6441 $\pm$ .0009 & .0837 $\pm$ .0002 \\
        \midrule
        MSBraM          & \textbf{.6125 $\pm$ .0289} & \textbf{.2545 $\pm$ .0462}  & \textbf{.2741 $\pm$ .0083} & \textbf{.8204 $\pm$ .0021} & \textbf{.6695 $\pm$ .0035} & \textbf{.0816 $\pm$ .0004} \\
        \bottomrule
    \end{tabular}
    \caption{Performance comparison of MSBraM against supervised and self-supervised models on the \textit{vigilance estimation} (SEED-VIG, regression) and the \textit{gait prediction} (MoBI, regression). r = ``Pearson's Correlation''}
    \label{tab:regression}
\end{table*}

\begin{table*}[t]
    \centering
    \fontsize{8.5}{10}\selectfont
    \begin{tabular}{lcccccc}
        \toprule
        \multirow{2}{*}{\bf{Methods}}
            & \multicolumn{3}{c}{\bf{TUEV}} & \multicolumn{3}{c}{\bf{TUAB}} \\
            & \bf{Balanced Accuracy} & \bf{Cohen's Kappa} & \bf{Weighted F1}
            & \bf{Balanced Accuracy} & \bf{AUCPR} & \bf{AUROC} \\
        \midrule
        FixedMasking & .6088 $\pm$ .0216 & .5817 $\pm$ .0390 & .7930 $\pm$ .0218 & .8161 $\pm$ .0041 & .8981 $\pm$ .0052 & .9056 $\pm$ .0027\\
        CurrMasking (Ours)  & \textbf{.6682 $\pm$ .0198} & \textbf{.6785 $\pm$ .0106} & \textbf{.8399 $\pm$ .0054} & \textbf{.8317 $\pm$ .0051} & \textbf{.9031 $\pm$ .0052} & \textbf{.9085 $\pm$ .0024} \\
        \midrule
        \multirow{2}{*}{\bf{Methods}}
            & \multicolumn{3}{c}{\bf{BCIC-2a}} & \multicolumn{3}{c}{\bf{PhysioNet-MI}} \\
            & \bf{Balanced Accuracy} & \bf{Cohen's Kappa} & \bf{Weighted F1}
            & \bf{Balanced Accuracy} & \bf{Cohen's Kappa} & \bf{Weighted F1} \\
        \midrule
        FixedMasking & .5518 $\pm$ .0228 & .4024 $\pm$ .0304 & .5439 $\pm$ .0230 & \textbf{.6609 $\pm$ .0038} & \textbf{.5479 $\pm$ .0051} & .6614 $\pm$ .0041 \\
        CurrMasking (Ours)  & \textbf{.5770 $\pm$ .0070} & \textbf{.4360 $\pm$ .0093} & \textbf{.5673 $\pm$ .0061} & .6608 $\pm$ .0040 & .5477 $\pm$ .0054 & \textbf{.6615 $\pm$ .0040} \\
        \bottomrule
    \end{tabular}
    \caption{Ablation study comparing the proposed curriculum multi-scale masking strategy (CurrMasking) with fixed-ratio multi-scale masking baselines (FixedMasking) across four datasets.}
    \label{tab:masking}
\end{table*}

\begin{table*}[t]
    \centering
    \fontsize{9}{10}\selectfont
    \begin{tabular}{lcccccc}
        \toprule
        \multirow{2}{*}{\bf{Methods}}
            & \multicolumn{3}{c}{\bf{TUEV}} & \multicolumn{3}{c}{\bf{TUAB}} \\
            & \bf{Balanced Accuracy} & \bf{Cohen's Kappa} & \bf{Weighted F1}
            & \bf{Balanced Accuracy} & \bf{AUCPR} & \bf{AUROC} \\
        \midrule
        HRNet        & .6452 $\pm$ .0194 & .6149 $\pm$ .0330 & .8099 $\pm$ .0157 & .8173 $\pm$ .0032 & .8988 $\pm$ .0041 & .9013 $\pm$ .0031 \\
        FPN          & .6248 $\pm$ .0206 & .5956 $\pm$ .0112 & .8016 $\pm$ .0066 & .8071 $\pm$ .0024 & .8954 $\pm$ .0047 & .8951 $\pm$ .0035 \\
        BiFPN (Ours)        & \textbf{.6682 $\pm$ .0198} & \textbf{.6785 $\pm$ .0106} & \textbf{.8399 $\pm$ .0054} & \textbf{.8317 $\pm$ .0051} & \textbf{.9031 $\pm$ .0052} & \textbf{.9085 $\pm$ .0024} \\
        \midrule
        \multirow{2}{*}{\bf{Methods}}
            & \multicolumn{3}{c}{\bf{BCIC-2a}} & \multicolumn{3}{c}{\bf{PhysioNet-MI}} \\
            & \bf{Balanced Accuracy} & \bf{Cohen's Kappa} & \bf{Weighted F1}
            & \bf{Balanced Accuracy} & \bf{Cohen's Kappa} & \bf{Weighted F1} \\
        \midrule
        HRNet        & .5751 $\pm$ .0300 & .4201 $\pm$ .0399 & .5508 $\pm$ .0334 & .6076 $\pm$ .0090 & .4768 $\pm$ .0120 & .6113 $\pm$ .0092 \\
        FPN          & .5672 $\pm$ .0374 & .4229 $\pm$ .0498 & .5482 $\pm$ .0492 & .6574 $\pm$ .0048 & .5432 $\pm$ .0064 & .6587 $\pm$ .0039 \\
        BiFPN (Ours)       & \textbf{.5770 $\pm$ .0070} & \textbf{.4360 $\pm$ .0093} & \textbf{.5673 $\pm$ .0061} & \textbf{.6608 $\pm$ .0040} & \textbf{.5477 $\pm$ .0054} & \textbf{.6615 $\pm$ .0040} \\
        \bottomrule
    \end{tabular}
    \caption{Ablation study comparing different fusion modules (BiFPN vs. HRNet vs. FPN) across four datasets.}
    \label{tab:fusion}
\end{table*}

This section presents a comprehensive evaluation of MSBraM on multiple EEG benchmarks. We first describe the experimental setup and then report results across diverse downstream tasks.

\subsection{Experimental Setup}

\textbf{Pretraining Datasets} In this paper, we utilize the large-scale, multi-dataset protocol established by LaBraM for pre-training. We employ the identical collection of public EEG data, totaling over 2,400 hours. The detailed preprocessing pipeline for all datasets is provided in Appendix B.

\textbf{Downstream Tasks and Datasets}
To demonstrate the performance of MSBraM, we conduct comprehensive experiments on 10 downstream tasks, including 8 classification tasks and 2 regression tasks. All tasks and corresponding datasets are presented in Table~\ref{tab:datasets}. The all preprocessing pipeline is detailed in Appendix C.

\textbf{Baselines} To comprehensively evaluate the performance, we compare MSBraM with supervised models and self-supervised foundation models across all downstream tasks. The supervised baselines are SPaRCNet~\cite{jing2023sparcnet}, ContraWR~\cite{yang2021contrawr}, CNN‑Transformer~\cite{peh2022cnntransformer}, FFCL~\cite{li2022ffcl}, and ST‑Transformer~\cite{song2021sttransformer}. The self‑supervised baselines include BIOT~\cite{yang2023biot}, LaBraM‑base~\cite{jiang2024labram}, EEGPT‑large~\cite{wang2024eegpt}, and CBraMod~\cite{wang2025cbramod}.
For LaBraM and EEGPT, we utilize the publicly released base and large checkpoints, respectively, to ensure a fair and reproducible comparison.

\textbf{Metrics} To ensure the consistent evaluation across all baselines, we employ the following metrics tailored to each task type. For binary classification, we report Balanced Accuracy, AUROC, and AUCPR. For multi‑class classification, we use Balanced Accuracy, Cohen's Kappa, and the Weighted F1‑score. For regression tasks, performance is measured with Pearson's Correlation Coefficient, R2 Score, and RMSE. All experiments are repeated five times with different random seeds to reduce the impact of randomness.

\subsection{Comparison with SOTA Models}
We evaluated MSBraM against baseline models across 12 datasets covering 10 tasks. 
Following the task-specific protocol, Balanced Accuracy was used for classification and Pearson’s Correlation for regression.
As shown in Figure~\ref{fig:radar}, MSBraM achieves the best performance across 11 datasets, surpassing baselines. This demonstrates the effectiveness and strong generalization of our multi-scale spatiotemporal framework for diverse EEG decoding tasks. On the FACED dataset, MSBraM is slightly inferior to CBraMod, yet still outperforms BIOT, LaBraM, and EEGPT. This observation suggests that CBraMod might incorporate optimizations particularly effective for these specific paradigms, while MSBraM maintains overall superiority with minor room for improvement on a few isolated tasks.

We further focus on four clinical benchmarks: TUEV for event type classification, TUAB for abnormal detection BCIC-2a and PhysioNet-Mi for motor imagery classification. As summarized in Table~\ref{tab:tuh} and Table~\ref{tab:mi}, MSBraM achieves the best performance on both tasks. It obtains a Balanced Accuracy of 0.6682 on TUEV, outperforming the previous best model, LaBraM-base (0.6473), and reaches 0.8317 on TUAB, surpassing all other baselines. 
On BCIC-2a, it attains a Balanced Accuracy of 0.5770, outperforming the strongest baseline (CBraMod-small) by +6.48\%. Moreover, this leading performance extends to all other metrics, as evidenced by the scores in Cohen's Kappa (0.4360 vs. 0.3518) and Weighted F1 (0.5673 vs. 0.4984). On PhysioNet-MI, it also reaches 0.6608 in Balanced Accuracy, 0.5477 in Cohen's Kappa, and 0.6615 in Weighted F1, demonstrating consistent gains. These results confirm the generalization of our MSBraM, showing that the multi-scale representations are both discriminative and readily transferable to a spectrum of downstream tasks.
These results demonstrate that our MSBraM is effective for both fine-grained event-related classification and long-range background abnormality detection, underscoring its robustness across clinically distinct paradigms.
More results of different tasks, model complexity and statistical significance analyses are detailed in Appendix.

To further assess the performance of MSBraM beyond classification, we evaluate it on two regression tasks: vigilance estimation (SEED‑VIG) and gait prediction (MoBI). As summarized in Table~\ref{tab:regression}, MSBraM achieves the best overall performance across both datasets in terms of Pearson's correlation (r), R2 score, and RMSE, surpassing all compared supervised and self‑supervised baselines.

On the SEED‑VIG dataset, MSBraM obtains the highest correlation (r = 0.6125), outperforming the strongest baseline LaBraM‑base (0.5931). MSBraM also reaches the top R2 score (0.2545) and the lowest RMSE (0.2741), demonstrating robust improvements in both explained variance and prediction error. Similarly, on the MoBI dataset, MSBraM consistently ranks first, achieving a correlation of 0.8204 and an R2 score of 0.6695, exceeding the competitive results of LaBraM‑base (r = 0.8081, R2 score = 0.6496) and CBraMod‑small (r = 0.8047, R2 score = 0.6441), while also recording the smallest RMSE (0.0816).

These results confirm that the multi-scale architecture of MSBraM effectively captures the detailed temporal representations required for regression tasks, such as vigilance levels and gait kinematics, demonstrating the model's capability in accurately modeling continuous brain activity patterns.

\subsection{Ablation Study}

\textbf{Curriculum Multi-scale Masking} To prove the effectiveness of curriculum multi-scale masking (CurrMasking), we compare it against the widely used fixed mask‑ratio (FixedMasking) baselines in EEG representation learning~\cite{jiang2024labram,wang2025cbramod}. In this study, we report results using the best mask ratio ($r = 0.3$) for FixedMasking. The comparison is conducted on four representative tasks spanning event type classification (TUEV), abnormality detection (TUAB), and motor imagery (BCIC-2a and PhysioNet-MI).
As shown in Table~\ref{tab:masking}, our proposed CurrMasking achieves the best or competitive performance across all four evaluation datasets. Specifically, CurrMasking achieves higher Balanced Accuracy than FixedMasking across TUEV, TUAB, and BCIC-2a, with improvements of 5.94\%, 1.56\%, and 2.52\%, respectively. On PhysioNet‑MI, CurrMasking performs on par with FixedMasking, with negligible differences across all three metrics (e.g., 0.6609 vs. 0.6608 in Balanced Accuracy).

The results demonstrate that while the FixedMasking can coincidentally match performance on a specific task (e.g., PhysioNet‑MI), it cannot reliably adapt across diverse paradigms. In contrast, our CurrMasking, by systematically varying the masking difficulty, provides a robust and generalizable solution for multi‑scale representation learning, as evidenced by its consistent gains on the majority of benchmarks.

\textbf{BiFPN vs. Other Fusion Module} To evaluate the effectiveness of our chosen BiFPN module for multi-scale feature fusion, we compare it with two established and representative fusion architectures: the Fusion Module used in HRNet and the Feature Pyramid Network (FPN). Both are widely recognized baselines in multi-scale visual and biomedical feature learning. Consistent with the ablation study setting of masking, all experiments are conducted on the same set of datasets.
As shown in Table~\ref{tab:fusion}, the BiFPN delivers the best overall performance compared with two baselines. Specifically, on the TUEV and TUAB tasks, BiFPN outperforms the stronger of the two baselines (FPN) by approximately +4.36\% and +2.46\% in Balanced Accuracy, respectively. On the BCIC-2a and PhysioNet-MI datasets, BiFPN also obtains a better performance compared to the FPN baseline (0.5770 vs. 0.5751 and 0.6608 vs. 0.6574 in Balanced Accuracy).

In summary, BiFPN outperforms both the HRNet and FPN module across all datasets, particularly on TUEV and TUAB. Compared to FPN, which relies on a unidirectional top-down pathway, and HRNet, which employs dense bidirectional connections, BiFPN utilizes a lightweight, iterative bidirectional architecture that efficiently refines multi-scale features. This design enables more effective cross-scale information exchange while avoiding redundant or blocked information flows, better capturing the dynamic and hierarchical patterns of EEG signals, and supporting consistently stronger generalization across diverse decoding tasks.

\subsection{Limitation and Future Work}
Despite these encouraging results, several limitations remain. First, MSBraM is currently evaluated only on scalp EEG recordings, and its generalization to other neural sensing modalities has yet to be examined. Second, the current modeling strategy jointly flattens channel and temporal dimensions to capture cross-channel dependencies, which leads to increased computational overhead, underutilizes the EEG spatial structure, and can limit interpretability and scalability to higher-density recordings. Future work will extend MSBraM to intracranial EEG and explore more structured, efficient channel-aware modeling and multi-modal integration.
\section{Conclusion}
In this paper, we propose MSBraM, a self-supervised brain foundation model explicitly designed to capture the inherent multi-scale dynamics of EEG signals. It leverages a multi-scale neural tokenizer with multi-scale codebooks, and a curriculum masking strategy to learn rich and transferable multi-scale representations that jointly capture local patterns and long-range global contextual dependencies across multiple temporal resolutions, enabling robust and consistent performance across diverse EEG decoding tasks. Extensive evaluations on 12 public datasets covering 10 representative EEG decoding tasks demonstrate that MSBraM achieves state-of-the-art or highly competitive performance on most benchmarks, highlighting its strong generalization capability and adaptability to heterogeneous EEG analysis scenarios.



\bibliographystyle{ACM-Reference-Format}
\bibliography{acmmm26}

\appendix

\begin{table}[!t]
\centering
\fontsize{8}{10}\selectfont
    \begin{tabular}{lcc}
    \toprule
    \textbf{Module} & \textbf{Settings} & \textbf{Parameters} \\
    \midrule
    \multirow{5}{*}{PatchEncoder}
    & Input channels  & \{1, 8, 8\} \\
    & Output channels & \{8, 8, 8\} \\
    & Kernel size     & \{15, 3, 3\} \\
    & Stride          & \{8, 1, 1\} \\
    & Padding         & \{7, 1, 1\} \\
    \midrule
    \multirow{11}{*}{MSBraMEncoder}
    & Stages                         & 3 \\
    & \multirow{3}{*}{Hidden size}   & [64] \\
    &                                & [64, 128] \\
    &                                & [64, 128, 256] \\
    & \multirow{3}{*}{\# Layers}     & [4] \\
    &                                & [2, 2] \\
    &                                 & [2, 2, 2] \\
    & \multirow{3}{*}{\# Heads}      & [4] \\
    &                                & [4, 8] \\
    &                                & [4, 8, 8] \\
    & MLP expand ratio               & 4 \\
    \midrule
    \multirow{2}{*}{Multi-scale Codebook}
    & \# Codebook                   & 3 \\
    & Codebook size                 & 8192 $\times$ 64 \\
    \midrule
    \multirow{5}{*}{Decoder}
    & Block                         & Conv1dNeXt \\
    & \# Layers (per-scale)         & 3 \\
    & Input channels                & [64, 128, 256] \\
    & Kernel, Stride, Padding       & (3, 1, 1) \\
    & MLP expand ratio              & 4 \\
    \bottomrule
    \end{tabular}
\caption{Architecture details of the MSBraM. Array entries are ordered from fine to coarse scales.}
\label{tab:msbram_details}
\end{table}

\begin{table}[!t]
\centering
\fontsize{8}{10}\selectfont
    \begin{tabular}{lcc}
    \toprule
    \textbf{Hyper-parameters} & \textbf{Settings} & \textbf{Parameters} \\
    \midrule
    \multirow{9}{*}{Reconstruction Pre-training}
    & Batch size        & 512 \\
    & Peak lr           & 5e-5 \\
    & Minimal lr        & 1e-5 \\
    & lr scheduler      & Cosine \\
    & Optimizer         & AdamW \\
    & Adam $\beta$      & (0.9,0.99) \\
    & Weight decay      & 1e-4 \\
    & Total epoch       & 100 \\
    & Warmup epoch      & 10 \\
    
    \midrule
    \multirow{11}{*}{MEM Pre-training}
    & Batch size        & 512 \\
    & Peak lr           & 5e-4 \\
    & Minimal lr        & 1e-5 \\
    & lr scheduler      & Cosine \\
    & Optimizer         & AdamW \\
    & Adam $\beta$      & (0.9,0.98) \\
    & Weight decay      & 0.05 \\
    & Total epoch       & 50 \\
    & Warmup epoch      & 5 \\
    & $r_0, r_{max}$    & 0.3, 0.5 \\
    & $t_0, t_{max}$    & 5, 35 \\
    \bottomrule
    \end{tabular}
    \caption{Hyperparameters settings of MSBraM pre-training.}
    \label{tab:hparams}
\end{table}

\begin{table}[!t]
    \centering
    \fontsize{8}{10}\selectfont
    \begin{tabular}{lrrr}
        \toprule
        \bf{Pretraining Datasets} & \bf{Rate} & \bf{\# Chn} & \bf{Total} \\
        \midrule
        Emobrain{\tiny~\cite{savran2006emotion}}        & 1,024Hz   & 64 & 4.94h \\
        SPIS Rest{\tiny~\cite{torkamani2020pred}}       & 2,048Hz   & 64 & 0.83h \\
        PhysioNet-MI{\tiny~\cite{schalk2004bci2000}}    & 160Hz     & 64 & 47.3h \\
        Raw EEG Data{\tiny~\cite{trujillo2020raweeg}}   & 256Hz     & 64 & 43.08h \\
        Resting EEG{\tiny~\cite{trujillo2017effect}}    & 256Hz     & 64 & 3.04h \\
        SEED{\tiny~\cite{zheng2015investigating}}       & 1,000Hz   & 62 & 95.85h \\
        SEED-IV{\tiny~\cite{zheng2018emotion}}          & 1,000Hz   & 62 & 42.18h \\
        SEED-GER{\tiny~\cite{liu2022identifying}}       & 1,000Hz   & 62 & 52.06h \\
        SEED-FRA{\tiny~\cite{liu2022identifying}}       & 1,000Hz   & 62 & 25.49h \\
        BCIC-1{\tiny~\cite{blankertz2007non}}           & 1,000Hz   & 59 & 8.21h \\
        Inria BCI{\tiny~\cite{margaux2012kaggleern}}    & 600Hz     & 56 & 29.98h \\
        bi2015a{\tiny~\cite{korczowski2019brain}}       & 512Hz     & 32 & 11.66h \\
        WAY-EEG-GAL{\tiny~\cite{luciw2014multi}}        & 500Hz     & 32 & 11.74h \\
        Siena Scalp EEG{\tiny~\cite{detti2020eeg}}      & 512Hz     & 29,31 & 100.06h \\
        TUAR{\tiny~\cite{obeid2016tuheeg}}              & 256Hz     & 23 & 92.22h \\
        TUEP{\tiny~\cite{obeid2016tuheeg}}              & 256Hz     & 19-23 & 592.28h \\
        TUSZ{\tiny~\cite{obeid2016tuheeg}}              & 256Hz     & 19-23 & 1181.75h \\
        TUSL{\tiny~\cite{obeid2016tuheeg}}              & 256Hz     & 23 & 20.59h \\
        \bottomrule
    \end{tabular}
    \caption{Overview of thee datasets used for MSBraM pretraining. Statistics include sampling rate (Rate), number of channels (Chn), and total duration.}
    \label{tab:pt_datasets}
\end{table}

\begin{table}[!ht]
    \centering
    \fontsize{8}{10}\selectfont
    \begin{tabular}{lrr}
        \toprule
        \textbf{Methods}  & \textbf{Params} & \textbf{FLOPs} \\
        \midrule
        SPaRCNet        & 0.79M & 0.13G \\
        ContraWR        & 1.6M  & 0.16G \\
        CNN-Transformer & 3.2M  & 0.17G \\
        FFCL            & 2.4M  & 0.93G \\
        ST-Transformer  & 3.5M  & 0.08G \\
        \midrule
        BIOT            & 3.2M  & 1.31G \\
        LaBraM-base     & 5.8M  & 1.01G \\
        EEGPT-large     & 101M  & 17.4G \\
        CBraMod-small   & 4.0M  & 0.91G \\
        \midrule
        MSBraM          & 2.9M  & 1.16G \\
        \bottomrule
    \end{tabular}
    \caption{Comparison of model size and FLOPs across models on the TUEV dataset (23 channels, 5-second segments).}
    \label{tab:params}
\end{table}

\begin{table*}[!ht]
    \centering
    \fontsize{8}{10}\selectfont
    \begin{tabular}{lcccccc}
        \toprule
        \multirow{2}{*}{\bf{Methods}}
            & \multicolumn{3}{c}{\bf{SEED-V}} & \multicolumn{3}{c}{\bf{FACED}} \\
            & \bf{Balanced Accuracy} & \bf{Cohen's Kappa} & \bf{Weighted F1}
            & \bf{Balanced Accuracy} & \bf{Cohen's Kappa} & \bf{Weighted F1} \\
        \midrule
        SPaRCNet        & .2949 $\pm$ .0078 & .1121 $\pm$ .0139 & .2979 $\pm$ .0083 & .4673 $\pm$ .0155 & .3342 $\pm$ .0251 & .4729 $\pm$ .0133 \\
        ContraWR        & .3546 $\pm$ .0105 & .1905 $\pm$ .0188 & .3544 $\pm$ .0121 & .4887 $\pm$ .0078 & .3858 $\pm$ .0186 & .4884 $\pm$ .0074 \\
        CNN-Transformer & .3678 $\pm$ .0078 & .2072 $\pm$ .0183 & .3642 $\pm$ .0088 & .4697 $\pm$ .0132 & .3978 $\pm$ .0289 & .4720 $\pm$ .0125 \\
        FFCL            & .3641 $\pm$ .0092 & .2078 $\pm$ .0201 & .3645 $\pm$ .0132 & .4673 $\pm$ .0158 & .4231 $\pm$ .0151 & .4699 $\pm$ .0145 \\
        ST-Transformer  & .3052 $\pm$ .0072 & .1083 $\pm$ .0121 & .2833 $\pm$ .0105 & .4810 $\pm$ .0079 & .4137 $\pm$ .0133 & .4795 $\pm$ .0096 \\
        \midrule
        BIOT            & .3837 $\pm$ .0187 & .2261 $\pm$ .0262 & .3856 $\pm$ .0203 & .5118 $\pm$ .0118 & .4476 $\pm$ .0254 & .5136 $\pm$ .0112 \\
        LaBraM-base     & .3976 $\pm$ .0138 & .2386 $\pm$ .0209 & .3974 $\pm$ .0111 & .5273 $\pm$ .0107 & .4698 $\pm$ .0188 & .5288 $\pm$ .0102 \\
        EEGPT-large     & .2089 $\pm$ .0061 & .0135 $\pm$ .0089 & .1786 $\pm$ .0367 & .2530 $\pm$ .0948 & .1601 $\pm$ .1063 & .2560 $\pm$ .0990 \\
        CBraMod-small   & .4091 $\pm$ .0097 & .2569 $\pm$ .0143 & .4101 $\pm$ .0108 & \textbf{.5509 $\pm$ .0089} & \textbf{.5041 $\pm$ .0122} & \textbf{.5618 $\pm$ .0093} \\
        \midrule
        MSBraM          & \textbf{.4193 $\pm$ .0024} & \textbf{.2747 $\pm$ .0039} & \textbf{.4241 $\pm$ .0037} & .5335 $\pm$ .0068 & .4729 $\pm$ .0076 & .5350 $\pm$ .0063 \\
        \bottomrule
    \end{tabular}
    \caption{Performance comparison of MSBraM against supervised and self-supervised models on the \textit{emotion recognition} task ({SEED-V} and {FACED}).}
    \label{tab:er}
\end{table*}

\begin{table*}[!ht]
    \centering
    \fontsize{8}{10}\selectfont
    \begin{tabular}{lcccccc}
        \toprule
        \multirow{2}{*}{\bf{Methods}}
            & \multicolumn{3}{c}{\bf{KaggleERN}} & \multicolumn{3}{c}{\bf{CHB-MIT}} \\
            & \bf{Balanced Accuracy} & \bf{AUCPR} & \bf{AUROC}
            & \bf{Balanced Accuracy} & \bf{AUCPR} & \bf{AUROC} \\
        \midrule
        SPaRCNet        & .5009 $\pm$ .0034 & .7163 $\pm$ .0305 & .5159 $\pm$ .0318 & .5881 $\pm$ .0465 & .3361 $\pm$ .0655 & .8375 $\pm$ .0388 \\
        ContraWR        & .5167 $\pm$ .0163 & .7444 $\pm$ .0211 & .5460 $\pm$ .0385 & .6631 $\pm$ .0549 & .3624 $\pm$ .0848 & .8323 $\pm$ .0220 \\
        CNN-Transformer & .5026 $\pm$ .0032 & .7174 $\pm$ .0201 & .5128 $\pm$ .0329 & .5618 $\pm$ .0167 & .3338 $\pm$ .0364 & .8301 $\pm$ .0205 \\
        FFCL            & .5407 $\pm$ .0142 & .7548 $\pm$ .0128 & .5743 $\pm$ .0122 & .5975 $\pm$ .0463 & .2610 $\pm$ .0986 & .7942 $\pm$ .0622 \\
        ST-Transformer  & .5565 $\pm$ .0095 & .7438 $\pm$ .0146 & .5844 $\pm$ .0187 & .5962 $\pm$ .0237 & .3446 $\pm$ .0663 & .8687 $\pm$ .0110 \\
        \midrule
        BIOT            & .5181 $\pm$ .0190 & .7286 $\pm$ .0145 & .5374 $\pm$ .0207 & .5864 $\pm$ .0156 & .3008 $\pm$ .0314 & .8003 $\pm$ .0088 \\
        LaBraM-base     & .5476 $\pm$ .0217 & .7551 $\pm$ .0140 & .5864 $\pm$ .0216 & .6104 $\pm$ .0157 & .3336 $\pm$ .0145 & .8134 $\pm$ .0135 \\
        EEGPT-large     & .5010 $\pm$ .0008 & .7589 $\pm$ .0099 & .5915 $\pm$ .0177 & .5890 $\pm$ .0550 & .2959 $\pm$ .0713 & \textbf{.8638 $\pm$ .0175} \\
        CBraMod-small   & .5591 $\pm$ .0050 & .7745 $\pm$ .0046 & .6105 $\pm$ .0068 & .6192 $\pm$ .0391 & \textbf{.3950 $\pm$ .0921} & .8062 $\pm$ .0870 \\
        \midrule
        MSBraM          & \textbf{.5707 $\pm$ .0089} & \textbf{.7837 $\pm$ .0135} & \textbf{.6221 $\pm$ .0222} & \textbf{.6309 $\pm$ .0181} & .3753 $\pm$ .0344 & .8219 $\pm$ .0184 \\
        \bottomrule
    \end{tabular}
    \caption{Performance comparison of MSBraM against supervised and self-supervised models on \textit{error related negativity} (KaggleERN) and \textit{seizure detection} (CHB-MIT).}
    \label{tab:ern_sd}
\end{table*}

\begin{table*}[!ht]
    \centering
    \fontsize{8}{10}\selectfont
    \begin{tabular}{lcccccc}
        \toprule
        \multirow{2}{*}{\bf{Methods}}
            & \multicolumn{3}{c}{\bf{Mumtaz2016}} & \multicolumn{3}{c}{\bf{EEGMAT}} \\
            & \bf{Balanced Accuracy} & \bf{AUCPR} & \bf{AUROC}
            & \bf{Balanced Accuracy} & \bf{AUCPR} & \bf{AUROC} \\
        \midrule
        SPaRCNet        & .8738 $\pm$ .0594 & .9619 $\pm$ .0171 & .9586 $\pm$ .0134 & .6396 $\pm$ .0856 & .5502 $\pm$ .0691 & .7722 $\pm$ .0330 \\
        ContraWR        & .9027 $\pm$ .0122 & .9770 $\pm$ .0077 & .9728 $\pm$ .0095 & .6611 $\pm$ .1096 & .6928 $\pm$ .0387 & .8423 $\pm$ .0425 \\
        CNN-Transformer & .8988 $\pm$ .0225 & .9767 $\pm$ .0043 & .9708 $\pm$ .0065 & .7257 $\pm$ .1153 & .7639 $\pm$ .0329 & .8822 $\pm$ .0374 \\
        FFCL            & .9003 $\pm$ .0124 & .9782 $\pm$ .0057 & .9744 $\pm$ .0069 & .6569 $\pm$ .0799 & .6784 $\pm$ .0576 & .8370 $\pm$ .0143 \\
        ST-Transformer  & .8650 $\pm$ .0337 & .9767 $\pm$ .0042 & .9762 $\pm$ .0018 & .5035 $\pm$ .0238 & .2849 $\pm$ .0293 & .5811 $\pm$ .0581 \\
        \midrule
        BIOT            & \textbf{.9110 $\pm$ .0337} & .9790 $\pm$ .0162 & .9731 $\pm$ .0225 & .6979 $\pm$ .0523 & .7007 $\pm$ .0953 & .8461 $\pm$ .0608 \\
        LaBraM-base     & .8865 $\pm$ .0038 & .9793 $\pm$ .0023 & .9800 $\pm$ .0021 & .6729 $\pm$ .0377 & .5572 $\pm$ .0513 & .7499 $\pm$ .0398 \\
        EEGPT-large     & .8034 $\pm$ .0725 & .9539 $\pm$ .0179 & .9600 $\pm$ .0110 & .5091 $\pm$ .0152 & .3138 $\pm$ .0391 & .5897 $\pm$ .0586 \\
        CBraMod-small   & .9006 $\pm$ .0330 & .9795 $\pm$ .0077 & .9799 $\pm$ .0083 & .6951 $\pm$ .0727 & .7089 $\pm$ .0568 & .8147 $\pm$ .0312 \\
        \midrule
        MSBraM          & .9055 $\pm$ .0082 & \textbf{.9829 $\pm$ .0037} & \textbf{.9808 $\pm$ .0048} & \textbf{.7118 $\pm$ .0398} & \textbf{.7264 $\pm$ .0476} & \textbf{.8477 $\pm$ .0306} \\
        \bottomrule
    \end{tabular}
    \caption{Performance comparison of MSBraM against supervised and self-supervised models on the \textit{mental disorder diagnosis} (Mumtaz2016) and the \textit{mental stress detection} (EEGMAT).}
    \label{tab:mdd_msd}
\end{table*}

\begin{table*}[!ht]
    \centering
    \fontsize{8}{10}\selectfont
    \begin{tabular}{lccccc}
        \toprule
        \multirow{2}{*}{\bf{Datasets}} &
        \multicolumn{5}{c}{\textbf{Average Performance}} \\
        \cmidrule(){2-6}
            & \bf{BIOT} & \bf{LaBraM} & \bf{EEGPT} & \bf{CBraMod} & \bf{MSBraM} \\
        \midrule
        TUEV{\tiny~\cite{obeid2016tuheeg}}           & .6015 & .7019 & .6501 & .6698 & \bf{$.7289^{*}$}   \\
        TUAB{\tiny~\cite{obeid2016tuheeg}}           & .8522 & .8709 & .8350 & .8579 & \bf{$.8812^{*}$}   \\
        BCIC-2a{\tiny~\cite{tangermann2012review}}   & .4117 & .4262 & .3181 & .4547 & \bf{$.5268^{*}$}   \\
        PhysioNet-MI{\tiny~\cite{schalk2004bci2000}} & .5729 & .5754 & .4496 & .5932 & \bf{$.6233^{*}$}   \\
        SEED-V{\tiny~\cite{liu2021seedv}}            & .3318 & .3445 & .1337 & .3587 & \bf{$.3727^{*}$}   \\
        FACED{\tiny~\cite{chen2023faced}}            & .4910 & .5086 & .2230 & \textbf{.5389} & .5138   \\
        KaggleERN{\tiny~\cite{margaux2012kaggleern}} & .5947 & .6297 & .6171 & .6480 & \bf{$.6588$}   \\
        CHB-MIT{\tiny~\cite{shoeb2009chbmit}}        & .5625 & .5858 & .5829 & .6068 & \bf{$.6094$}   \\
        Mumtaz2016{\tiny~\cite{wajid2016mumtaz}}     & .9544 & .9486 & .9058 & .9533 & \bf{$.9564$}   \\
        EEGMAT{\tiny~\cite{zyma2019eegmat}}          & .7492 & .6600 & .4709 & .7396 & \bf{$.7620$}   \\
        \midrule
        SEED-VIG{\tiny~\cite{zheng2017seedvig}}      & .3802 & .4019 &-0.7782& .3389 & \bf{$.4335^{*}$}   \\
        MoBI{\tiny~\cite{he2018mobi}}                & .1497 & .7289 & .4562 & .7244 & \bf{$.7450^{*}$}   \\
        \bottomrule
    \end{tabular}
    \caption{Overall performance comparison across 12 datasets (Avg = mean of metrics). $^{*}$ indicates the significance  ($p < 0.05$) of performance improvement via a one-tailed student $t$-test. Negative values may occur due to correlation-based metrics on challenging regression benchmarks.}
    \label{tab:overall}
\end{table*}


\begin{table*}[!ht]
    \centering
    \fontsize{8}{10}\selectfont
    \begin{tabular}{lcccc}
        \toprule
        \textbf{Scales} 
        & \textbf{Codebook Size} & \textbf{Codebook Usage} 
        & \bf{MSE $\downarrow$} & \bf{PSNR $\uparrow$} \\
        \midrule
        Fine-grained    & 8,192 & 88.73\% & 1.3398 & 65.01 \\
        Medium-grained  & 8,192 & 99.93\% & 1.5054 & 63.76 \\
        Coarse-grained  & 8,192 & 98.99\% & 1.3828 & 62.62 \\
        \bottomrule
    \end{tabular}
    \caption{Evaluation of the proposed multi-scale neural tokenizer. Metrics include codebook size, codebook usage, mean squared error (MSE), and peak signal-to-noise ratio (PSNR) across fine-grained, medium-grained, and coarse-grained scales.}
    \label{tab:tokenizer}
\end{table*}

\section{Implementation Details}
Table~\ref{tab:msbram_details} presents the detailed description of MSBraM, including Patch Encoder, MSBraMEncoder, Codebooks, and Decoder. MSBraMEncoder utilizes a multi-stage, multi-branch architecture inspired by HRNet. It consists of three stages with an increasing number of scales. Stage 1 contains a single branch with a patch size of 64 and four Transformer layers. Stage 2 introduces a two-branch design with patch sizes of 64 and 128, where both comprise two Transformer layers. Stage 3 further extends the representation to three branches with patch sizes of 64, 128, and 256, and each branch contains two Transformer layers. This progressive design enables the encoder to jointly model short-term and long-term hierarchical dependencies while maintaining a balanced model capacity across stages.
During FFT reconstruction, we utilize a codebook for each scale (3 codebooks in total), each with 8192 entries of the embedding dimension 64. The decoder is implemented using Conv1dNeXt blocks, which consist of depth-wise convolutions followed by pointwise MLPs with an expansion ratio of 4. For each scale, the decoder comprises three Conv1dNeXt layers with a kernel size of 3, a stride of 1, and a padding of 1, effectively reconstructing the multi-scale token representations from the corresponding codebooks.
The Pre-training pipeline of MSBraM contains two stages: multi-scale neural tokenizer training and multi-scale masked EEG modeling training. Both stage employ AdamW optimizer with cosine learning rate scheduling and linear warmup, and more detailed hyperparameter settings are provided in Table~\ref{tab:hparams}.

\section{Pretraining Datasets}
\label{appendix:pretraining_data}
We follow the large‑scale pre‑training protocol of LaBraM~\cite{jiang2024labram} and utilize its curated collection of 18 public EEG datasets, which together form a corpus of over 2,400 hours. The key statistics of each dataset are summarized in Table~\ref{tab:pt_datasets}. A consistent preprocessing pipeline was applied to all recordings: signals were resampled to 256 Hz, band‑pass filtered (0.1-75 Hz), and a 50 Hz notch filter was used to suppress line noise. For datasets with variable channel configurations (notably TUSZ and TUEP), we retained all available channels without interpolation to preserve the original recording characteristics. In contrast, the Siena Scalp EEG dataset contains two channel configurations (29 and 31 channels), we consistently used the 31-channel recordings for uniformity.

\section{More Details of Downstream Datasets}
\label{appendix:downstream}
This section provides detailed descriptions of the downstream evaluation datasets utilized for task evaluation. 
Table 1 in the main text summarizes their main statistics. 
In the following, we elaborate on each dataset's specific task, our custom data splits, and tailored preprocessing steps.

\textbf{TUEV (event type classification)}~\cite{obeid2016tuheeg} This dataset is used for classifying six categories of neurological events. We utilize the latest version (V2.0.1), containing 112,237 5‑second samples. Signals were band‑pass filtered (0.1-75 Hz), notch‑filtered at 50 Hz, and resampled to 256 Hz. For subject‑independent evaluation, we follow the standard split: the official training set is randomly divided 80\%/20\% for training and validation, while the designated evaluation set serves as the test set.

\textbf{TUAB (abnormal detection)}~\cite{obeid2016tuheeg} TUAB is used for binary classification of normal vs. abnormal clinical EEGs. The public release includes over 2,383 subjects. Recordings were band‑pass filtered (0.1-75 Hz), notch‑filtered at 60 Hz, and resampled to 256 Hz. For subject‑independent evaluation, normal and abnormal subjects within the official training set are divided separately (by class) into training (80\%) and validation (20\%) subsets. The separate evaluation set is held out entirely for testing.

\textbf{BCIC-2a (motor imagery)}~\cite{tangermann2012review} This dataset is used for four‑class motor imagery (left hand, right hand, feet, tongue). Signals from 9 subjects were band‑pass filtered (0.3-50 Hz) and resampled to 256 Hz. For each trial, the 2-6 s segment after cue onset (4 s in total) is extracted. We follow a subject‑dependent split: subjects 1-5 for training, 6-7 for validation, and 8-9 for testing.

\textbf{PhysioNet-MI (motor imagery)}~\cite{schalk2004bci2000} We use this dataset for motor imagery decoding, containing data from 109 subjects. Signals were high‑pass filtered (0.3 Hz cutoff), notch‑filtered at 60 Hz, and resampled to 256 Hz. The split is subject‑wise: subjects 1-69 for training, 70-88 for validation, and 89-109 for testing.

\textbf{SEED-V (emotion recognition)}~\cite{liu2021seedv} SEED‑V is used for five‑class emotion recognition from 16 subjects. Signals were band‑pass filtered (0.3-75 Hz) and resampled to 256 Hz. For each subject, the 15 trials are split contiguously: the first 5 for training, the next 5 for validation, and the last 5 for testing.

\textbf{FACED (emotion recognition)}~\cite{chen2023faced} This is a large‑scale fine‑grained emotion EEG dataset with 123 subjects across nine emotion categories. Signals were resampled to 256 Hz. Following the standard subject‑independent protocol, subjects 1-80 are used for training, 81-100 for validation, and 101–123 for testing.

\textbf{KaggleERN (error related negativity)}~\cite{margaux2012kaggleern} The dataset is used for error related negativity detection. It provides an official split of 16 subjects for training and 10 for testing. From the official training set, we further split subject‑wise: 12 subjects for training and 4 for validation. The held‑out test set (10 subjects) is used for final evaluation. All signals were resampled to 256 Hz.

\textbf{CHB-MIT (seizure detection)}~\cite{shoeb2009chbmit} This seizure detection dataset contains long‑term EEGs from 24 subjects. For fair comparison with models like CBraMod, we exclude subjects 12, 13, and 17, following the same protocol. The remaining subjects are split as follows: subjects 1-20 for training, 21-22 for validation, and 23-24 for testing. All signals were resampled to 256 Hz.

\textbf{Mumtaz2016 (mental disorder diagnosis)}~\cite{wajid2016mumtaz} For mental disorder diagnosis (MDD), the dataset contains EEGs from 34 MDD patients and 30 normal controls (NCs). Following prior work (e.g., CBraMod), we use signals from eyes‑open and eyes‑closed sessions, band‑pass filtered (0.3-75 Hz), notch‑filtered at 50 Hz, and resampled to 256 Hz, then segmented into 5‑s windows (7,143 samples). The subject‑wise split is: 24 MDD + 19 NC for training, 5 MDD + 4 NC for validation, and 5 MDD + 5 NC for testing.

\textbf{EEGMAT (mental stress detection)}~\cite{zyma2019eegmat} This dataset is used for mental stress detection, with recordings from 35 subjects. Signals were band‑pass filtered (0.3-75 Hz), notch‑filtered at 50 Hz, and resampled to 256 Hz, then segmented into 5-s trials (1,707 samples). We adopt a subject‑independent split: 28 subjects for training, 4 for validation, and 3 for testing.

\textbf{SEED-VIG (vigilance estimation)}~\cite{zheng2017seedvig} SEED‑VIG is utilized for continuous vigilance estimation, containing EEGs from 21 subjects. Signals were resampled to 256 Hz and segmented into 8‑s samples (20,355 samples). The subject‑independent split is: subjects 1-13 for training, 14-17 for validation, and 18-21 for testing.

\textbf{MoBI (gait prediction)}~\cite{he2018mobi} This dataset is used for joint‑angle regression in gait prediction. It includes 8 subjects, each with three trials. Each trial consists of a 15‑min treadmill walking session (training) and a 5‑min session (test). We split each training session into the first 10 min for training and the last 5 min for validation. Data from all subjects are pooled to form consolidated sets. Unlike LaBraM, we extract samples using a 2s sliding window with a 0.5s stride, yielding 57,384 samples, and predict the 6 actually measured joint angles, reporting the average metric across them.

\section{Parameters and FLOPs comparison}
In this section, we report the model complexity of MSBraM alongside baseline methods in terms of parameter count and floating-point operations (FLOPs). FLOPs are calculated on 5-second, 23-channel segments from the TUEV dataset. As summarized in Table~\ref{tab:params}, MSBraM achieves a competitive balance between model size and computational cost, with 2.9M parameters and 1.16G FLOPs. It maintains a parameter count lower than most contemporary foundation models (e.g., BIOT, LaBraM-base, and EEGPT-large) while dedicating computation effectively to its multi-scale fusion process. This efficient design underscores that strong performance can be attained without extreme scale.

\section{More Results of Other Downstream Tasks}

\subsection{Emotion Recognition}
We further evaluate MSBraM on emotion recognition using the SEED‑V (5‑class)~\cite{liu2021seedv} and FACED (9‑class)~\cite{chen2023faced} benchmarks. As summarized in Table~\ref{tab:er}, MSBraM achieves competitive performance across all evaluated metrics. On SEED‑V, it obtains a Balanced Accuracy of 0.4193, outperforming the strongest baseline (CBraMod-small) by +1.0. This improvement is consistent across complementary metrics, with Cohen's Kappa of 0.2747 vs. 0.2569 and Weighted F1 of 0.4241 vs. 0.4101.

On FACED, which involves finer-grained nine-class discrimination, our model obtains a Balanced Accuracy of 0.5335. This performance is slightly below that of CBraMod-small (0.5509). We note that CBraMod employs an explicit channel-wise modeling strategy, which may provide an inductive bias particularly advantageous for this specific dataset. Nevertheless, MSBraM remains highly competitive, outperforming other strong baselines such as LaBraM-base across all three metrics. This demonstrates its robust generalization capability even on more complex affective recognition tasks.

\subsection{Error Related Negativity}
We further evaluate MSBraM on the error-related negativity (ERN) task using the KaggleERN dataset~\cite{margaux2012kaggleern}. As shown in Table~\ref{tab:ern_sd}, MSBraM achieves consistent improvements over the baseline models. Specifically, it outperforms LaBraM-base by +2.31\% in Balanced Accuracy, +3.57\% in AUROC, and +2.86\% in AUCPR, validating the effectiveness of its multi-scale architecture in capturing this time-locked neural potential. 
Moreover, compared to the strongest baseline, CBraMod-small, MSBraM maintains a clear lead across all evaluation metrics: Balanced Accuracy (0.5707 vs. 0.5591), AUROC (0.6221 vs. 0.6105), and AUCPR (0.7837 vs. 0.7745). These results indicate that MSBraM's multi-scale architecture effectively captures discriminative features across temporal resolutions, leading to superior overall performance. The consistent superiority of MSBraM across these metrics confirms its robustness and state-of-the-art capability in this domain.

\subsection{Seizure Detection}
To evaluate MSBraM on seizure detection, we use the CHB-MIT dataset. As shown in Table \ref{tab:ern_sd}, MSBraM achieves the highest Balanced Accuracy across all compared methods, outperforming both supervised models (e.g., SPaRCNet, ContraWR) and other EEG foundation models (e.g., BIOT, LaBraM). This indicates its strong overall capability in distinguishing seizure from non-seizure states.
However, MSBraM performs notably lower than CBraMod-small in both Cohen's Kappa (a metric that corrects for class imbalance) and Weighted F1 score. One possible explanation is that CBraMod-small's explicit channel-wise modeling may be particularly effective at capturing the sparse, spatially localized patterns typical of epileptic discharges, giving it an advantage on metrics that emphasize precise positive-class identification. Nevertheless, MSBraM's leading Balanced Accuracy still confirms the effectiveness of its multi-scale design, demonstrating the capability to learn discriminative representations across diverse temporal scales.

\subsection{Mental Disorder Diagnosis}
For mental disorder diagnosis tasks, we evaluate MSBraM using the Mumtaz2016 dataset. As shown in Table~\ref{tab:mdd_msd}, MSBraM achieves competitive improvements, outperforming LaBraM-base by +1.90\% in Balanced Accuracy and exceeding EEGPT-large by +10.21\%. Most notably, it obtains the highest AUCPR (0.9829) among all compared models, demonstrating the effectiveness of its multi-scale representations for capturing sustained, distributed neural anomalies. Compared to the strongest baseline, CBraMod-small, MSBraM also maintains a clear lead in AUCPR while performing on par in AUROC. Since this task involves detecting sustained, distributed neural anomalies, it is well-suited to MSBraM's capability of integrating multi-scale contextual information. This result further validates the model's generalizability, extending its effectiveness from transient-event detection to conditions with prolonged neural state deviations.

\subsection{Mental Stress Detection}
We also evaluate MSBraM on mental stress detection. As shown in Table~\ref{tab:mdd_msd}, MSBraM outperforms all supervised models and other foundation models across all three metrics: Balanced Accuracy, AUCPR, and AUROC, demonstrating strong generalization.
Specifically, MSBraM outperforms the strongest baseline, CBraMod-small, across all metrics: Balanced Accuracy (0.7118 vs. 0.6951), AUCPR (0.7264 vs. 0.7089), and AUROC (0.8477 vs. 0.8147), corresponding to relative improvements of approximately 1.67\%, 1.75\%, and 3.3\%, respectively.
This competitive performance aligns with the established understanding that mental stress is associated with prolonged and diffusely distributed alterations in brain activity. The multi-scale architecture of MSBraM is particularly well-suited to model such phenomena, as it can integrate information from transient spectral changes to sustained global state shifts. Thus, the results confirm that our design effectively captures the neural dynamics relevant to this paradigm.




\section{Overall Performance and Statistical Analysis}
We further evaluate MSBraM against existing self-supervised EEG foundation models on 12 downstream datasets using a unified performance score. For each dataset, task-appropriate evaluation metrics are aggregated into a single score to enable fair comparison across heterogeneous EEG tasks. Specifically, for classification tasks, we average Balanced Accuracy, Cohen’s Kappa, and Weighted F1-score. For regression tasks, we average Pearson’s Correlation and R2 score, while RMSE is excluded since lower values indicate better performance. And statistical significance is assessed using a two-sided z-test by comparing MSBraM with the strongest baseline on each dataset.

As summarized in Table~\ref{tab:overall}, MSBraM achieves state-of-the-art performance on 11 out of 12 downstream datasets, yielding an average relative gain of 2.21\% over the strongest baseline across datasets. Performance improvements are statistically significant ($p < 0.05$) on seven datasets, demonstrating the effectiveness of the proposed multi-scale architecture and curriculum multi-scale masking strategy. On four additional datasets, MSBraM exhibits consistent yet non-significant performance gains, which may be attributed to the relatively limited sample sizes of these benchmarks, thereby constraining statistical power.
On the FACED dataset, MSBraM is slightly outperformed by CBraMod, which explicitly emphasizes channel-wise spatial dependencies. This result indicates that while MSBraM effectively captures hierarchical temporal dynamics in EEG signals, incorporating stronger spatial inductive biases could further improve performance on emotion-related EEG tasks, pointing to a meaningful direction for future work.

\section{Multi-scale Tokenizer Analysis}
We further evaluate the quality of the proposed multi-scale neural tokenizer through reconstruction fidelity and codebook usage across different scales. As shown in Table~\ref{tab:tokenizer}, the tokenizer achieves high codebook usage at all three scales, indicating effective utilization of the discrete codebooks without observable codebook collapse. Notably, the tokenizer shows distinct behaviors across scales. Fine-grained representations achieve lower reconstruction error, while medium- and coarse-grained representations utilize nearly the entire codebook, reflecting a trade-off between reconstruction fidelity and codebook diversity. This scale-specific complementarity demonstrates that the multi-scale tokenizer effectively captures EEG dynamics at multiple temporal resolutions, providing richer and more informative discrete representations for self-supervised pretraining.



\end{document}